\pdfoutput=1

\documentclass[sigconf]{aamas}
\renewcommand\footnotetextcopyrightpermission[1]{} 
\usepackage{booktabs}

\acmConference[Technical Report]{}

\begin{document}

\title{Recognizing Plans by Learning Embeddings from Observed Action Distributions}  




\author{Yantian Zha, Yikang Li, Sriram Gopalakrishnan, Baoxin Li, Subbarao Kambhampati}
\affiliation{%
  \institution{Arizona State University}
  \streetaddress{699 S Mill Ave}
  \city{Tempe} 
  \state{Arizona} 
  \postcode{85281}
}
\email{{Yantian.Zha, yikangli, sgopal28, baoxin.li, rao}@asu.edu}

\maketitle
\begin{abstract}  
Recent advances in visual activity recognition have raised the possibility of applications such as automated video surveillance. Effective approaches for such problems however require the ability to recognize the plans of agents from video information. Although traditional plan recognition algorithms depend on access to sophisticated planning domain models \cite{geffner-ramirez,sohrabi2016plan}, one recent promising direction involves learning approximated (or shallow) domain models directly from the observed activity sequences \cite{dup}. One limitation is that such approaches expect observed action sequences as inputs. In many cases involving vision/sensing from raw data, there is considerable uncertainty about the specific action at any given time point. The most we can expect in such cases is probabilistic information about the action at that point. The input will then be sequences of such observed action distributions. In this work, we address the problem of constructing an effective data-interface that allows a plan recognition module to directly handle such observation distributions. Such an interface works like a bridge between the low-level perception module, and the high-level plan recognition module. We propose two approaches. The first involves resampling the distribution sequences to single action sequences, from which we could learn an action affinity model based on learned action (word) embeddings for plan recognition. The second is to directly learn action distribution embeddings by our proposed {\Distr2vec} (distribution to vector) model, to construct an affinity model for plan recognition. 
\end{abstract}

%




\pdfoutput=1
\def\DUP{\tt DUP}
\def\UDUP{\tt UDUP}
\def\Distr2vec{\tt Distr2Vec}
\def\Word2vec{\tt Word2Vec}

\def\RBM{\tt RBM}
\def\NM{\tt NM}
\def\one{\mathbb{I}}

\section{Introduction}


Many applications like surveillance require plan recognition to predict future actions of an agent. The data to such a model-based plan recognition module would have to come from visual/sensory recognition module. Rather than provide a sequence of ground truth actions, such a visual recognition module would provide a sequence of distributions over actions. Each distribution (per step of the sequence) would capture the uncertainty in the action observed at that point. In this paper, our focus is on training shallow domain models for plan recognition, that can handle such observational uncertainty. 


\begin{figure*}[thpb] 
  \centering
  \includegraphics[width=12cm]{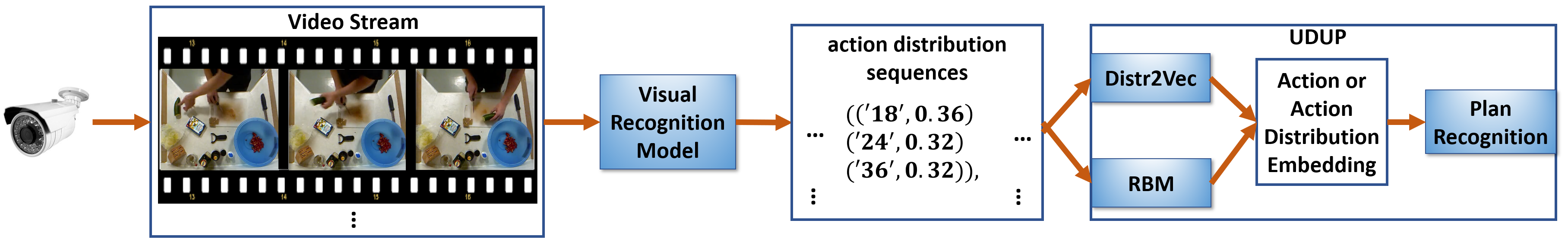}
  \caption{This figure illustrates the application of UDUP on uncertain observations from a visual recognition model.}
  \label{intro_udup}
\end{figure*}

One approach that could address plan recognition without uncertainty in action recognition, is {\DUP} \cite{dup}. In {\DUP} the approach taken was to learn a Skip-gram {\Word2vec} model \cite{word2vec} to get embeddings for the actions in the action sequences. {\Word2vec} was originally intended for learning embeddings for words in sentences. However, by treating actions as words, and plan traces as sentences, {\DUP} can learn embeddings for actions as well. Such embeddings capture the affinities of the actions, which is why {\DUP} used it for plan recognition. 

That being said, {\DUP} does not handle recognition uncertainty in visual outputs. In our framework {\UDUP} (Uncertain {\DUP}), we propose two approaches that can handle uncertainty and learn embeddings for plan recognition. One is Resampling-Based Model ({\RBM}) and the other is {\Distr2vec} model. The overall system is illustrated in Figure \ref{intro_udup}. {\RBM} samples action sequences from each distribution sequence, and then uses those (sampled) single action sequences as the training data for {\Word2vec} to learn embeddings. Our other approach, {\Distr2vec}, comes with a more fundamental change to {\Word2vec} that could learn action (word) distribution embeddings from distribution sequences. To train the {\Distr2vec}, we introduce a loss function based on combining KL-divergence and hierarchical softmax \cite{word2vec}. KL-divergence has been well known to be good at measuring the distance between two distributions, which we can use to update the model, while keeping the input data as is (no resampling). Thus, the overall idea is that the embeddings that can preserve the distribution information are a more informative data interface between the visual recognition module and the plan recognition module. We empirically demonstrate that our two approaches, {\RBM} and {\Distr2vec}, improve plan recognition in {\UDUP} over the baseline {\Word2vec} approach applied in {\UDUP}. We evaluate our approaches on both a synthetic plan corpora, and a plan corpora from real world videos. Although we developed and evaluated our approaches in the context of plan recognition, they will be useful in any categorical data sequence learning scenario where there is uncertainty at each step of input data. 

\section{Problem formulation} \label{prob-form}
The previous section explains how single action sequences could be used to learn an affinity model as an shallow domain model for plan recognition. In this section, we define our problem of learning shallow domain models from action distributions for plan recognition.

	The input is a plan trace library $L$. Each plan trace $p$ is a sequence of distributions of actions. We define the action space as $A=\bar{A} \cup {\phi}$. $A$ consists of all possible grounded action symbols ($\bar{A}$), and a symbol $\phi$ which denotes unknown missing action in a position. Thus, the plan recognition problem is defined as $R=(L,O,A)$, and $O$ are sequences of distributions of actions. The solution $R$ is either a complete plan with all missing actions filled in, or a plan that includes future actions of an agent.
    
In each step of the plan trace $p$, we use the $K$ probable actions as well as their confidences $c$ (equal to their probabilities) for each step in the trace, to produce a trace of  distributions.
    The confidence $c$ of an action is a value given by the low-level perception module, representing how confident or probable it is for that particular observation. If there are $T$ time-steps, and $K$ values for each distribution (in a time step), then an uncertain plan trace $p$ is represented as a matrix below.
 

\newcommand{\mymatrix}[1]{\ensuremath{\left\downarrow\vphantom{#1}\right.\overset{\xrightarrow[\hphantom{#1}]{\text{Time}}}{#1}}}
\begin{equation*}
\text{\tiny Actions}\mymatrix{\begin{pmatrix}
    a^1_1,c^1_1 & a^1_2,c^1_2 & a^1_3,c^1_3 & \cdots & a^1_T,c^1_T\\ 
    a^2_1,c^2_1 & a^2_2,c^2_2 & a^2_3,c^2_3 & \cdots & a^2_T,c^2_T\\  
    a^3_1,c^3_1 & a^3_2,c^3_2 & a^3_3,c^3_3 & \cdots & a^3_T,c^3_T\\ 
    \vdots & \vdots & \vdots & \ddots & \cdots \\
    a^K_1,c^K_1 & a^K_2,c^K_2 & a^K_3,c^K_3 & \cdots & a^K_T,c^K_T 
        \end{pmatrix}}
\end{equation*}
    
	We use $Distr(a_t)$ to denote the distribution $\langle (a^1_t,c^1_t),(a^2_t,c^2_t),...,$ $(a^K_t,c^K_t)\rangle$ at a specific time step. We can now formulate the task of training action embeddings and shallow domain models, as maximizing the log probability of distributions as follows:

\begin{equation}\label{udup_problem}
\frac{1}{T}\sum_{t=1}^T\sum_{-\mathcal{W}\leq j\leq \mathcal{W},j\neq0}\log p(Distr(a_{t+j})|Distr(a_t))
\end{equation} where $Distr(a_t)$ is the input observation distribution, $Distr(a_{t+j})$ is the target observation distribution, and $\mathcal{W}$ is the window size.  

\begin{figure}[!ht]
\begin{center}
	\centerline{\includegraphics[width=0.44\textwidth]{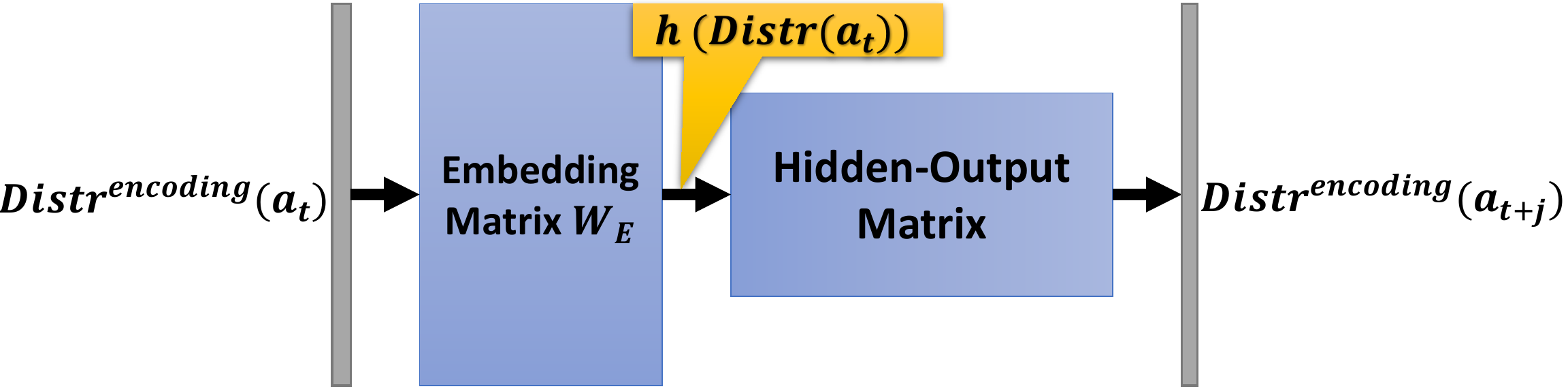}}
	\caption{The architecture of our {\Distr2vec} model for learning distribution embeddings and shallow planning domain models.}
	\label{m2_arch}
\end{center}
\end{figure}

Also, when feeding $Distr(a_t)$ into the input layer of {\Distr2vec} (the architecture is shown in Figure \ref{m2_arch}), we actually encode $Distr(a_t)$ into a vector $Distr^{encoding}(a_t)={x_1,...,x_{\bar{A}}}$. The size of $\bar{A}$ equals to the number of nodes in input layer. We encode $Distr(a_t)$ into a vector by having a unique index $i$ associated to each action in $\bar{A}$. The value $v_i$ at each index $i$ in the input vector, is the confidence value associated to the matching action in $Distr(a_t)$. The rest units of input layer, whose corresponding actions are not in $Distr(a_t)$, have zero probability values.

\section{UDUP Framework} \label{udup}
Our {\UDUP} framework (shown in the Figure \ref{intro_udup}) consists of two parts: learning a model that captures action distribution affinities as the shallow domain model (explained in Section \ref{rbm} and \ref{dist2vec-md}) from plan corpora (have complete plan traces), and recognizing plans by maximizing the affinities of actions in (originally incomplete) plan traces with filled actions (hence obtain a plan completion). Each input observation trace is read in as a $K \times T$ matrix, as explained in Section \ref{prob-form}. If the plan trace $p$ is incomplete, there are steps in which observed distributions are missing. All of these steps form an unknown plan $\widetilde{p}$, which requires actions from $\bar{A}$ to be filled in. Once all such steps with missing observations have a predicted action filled in, then the plan $p$ has been recognized.


If there are totally $M$ missing positions in an unknown plan $\widetilde{p}$ (only exists during testing or plan recognition phase), we try different actions to fill in each position in $\widetilde{p}$, and select the action that maximize the affinities. For example, consider the case that the trace we are testing on has only one missing observation (at the position $j$). All observations at other positions are known, and are action distributions. Then that single position $j$ forms an unknown plan $\widetilde{p}$. To determine the missing action, we calculate $\bar{A} \times 2\mathcal{W}$ pairwise affinities for all possible actions that could complete $\widetilde{p}$. $\mathcal{W}$ is the window size (a hyper\-parameter in {\Word2vec}) and is set to one. To score $\widetilde{p}$ with one filled-in action in our example, we need to measure the pairwise affinities between all possible actions in $\bar{A}$, and an observed action distribution in the context window $-\mathcal{W}\leq j \leq \mathcal{W}$. The selected action would have the highest pairwise affinity to observations in the window. As a result, because the plan only has one position with missing observation (length of $\widetilde{p}$ equals to one), that single action forms the $\widetilde{p}$ that has the highest score.

If there are $M$ ($M>1$) positions with missing observations (length of $\widetilde{p}$ is $M$), we iterate through each missing position in the following manner. First, for a missing position, we need to identify totally $\bar{A} \times 2\mathcal{W}$ pairs (set $\mathcal{W}$ to one). With each position $t$ in the trace as the center, we enumerate all pairs of an action (from $\bar{A}$ that may be filled in), and an observed action distribution in context window (i.e., those between $t-\mathcal{W}$ and $t+\mathcal{W}$). Based on learned embeddings we try different actions (from $\bar{A}$) and obtain corresponding pairwise affinity values. Once we finish the iteration, we predicted all $M$ actions and calculate a score for the current completion of $\widetilde{p}$. The completion that has the highest score would be the recognized plan.

The pairwise affinity is calculated using Equation \ref{final-m2} (implemented as a function $\textit{affinity\_pair}$ in Equation \ref{score}). We use Equation \ref{score} to compute the score $\mathcal{F}(\widetilde{p})$ of a plan completion for an unknown plan $\widetilde{p}$.



    
\begin{equation} \label{score}
\mathcal{F}(\widetilde{p})=\sum_{t=1}^M \sum_{-\mathcal{W}\leq j \leq \mathcal{W},j \neq 0} \textit{affinity\_pair}(input, target)
\end{equation} 
where both input and target are encoding of either a single action (a one-hot vector), enumerated from the action vocabulary, or an observed action distribution (e.g., $Distr^{encoding}(a_{t})$). They are the input for an aforementioned function $affinity\_pair$. $p$ is a plan with unknown observations. Each $\widetilde{p}$ (has length $M$) is a plan completion, in which the $M$ actions would be filled in $M$ positions with missing observations in an incomplete plan $p$. $j$ is the index inside a context window of size $\mathcal{W}$.

The plan completion for $\widetilde{p}$ that have the highest score, would be treated as the solution for the recognition of plan $p$.



\section{Resampling Based Model (RBM)} \label{rbm}
In {\RBM}, we first calculate the likelihoods of all possible paths from each trace of action distributions. The path weights ($PW$) can be calculated by multiplying the confidence values of all actions along a path, and thus could be used to represent the overall uncertainty of that path. The top $N$ action sequences with probabilities (ranked according to $PW$s) are selected, and then resampled using the roulette wheel resampling approach of \cite{lipowski2012roulette}. This lets us drop some sampled traces that have low probabilities, and increase the amount of samples with high probabilities. The set of traces selected after the resampling step is used to train a {\Word2vec} model. Readers may want to refer to Section \ref{wvhs} for technical details of the {\Word2vec} that we used.

That said, the potential problem for {\RBM} are the following. Firstly, resampling would make the algorithm more computationally expensive, and the training time would be closely related to the number of samples and length of the training data. This makes it hard to apply RBM to a real-time plan recognition system. Secondly, the resampling approach would lose some information. The solution would be to directly use the entire distribution sequence. This leads to our design of {\Distr2vec} model.


\section{Distr2Vec model} \label{dist2vec-md}

We learn shallow planning domain models from uncertain data (distributions), by maximizing Equation \ref{udup_problem}. 
We train our {\Distr2vec} model by minimizing the Kullback-Leibler (KL) divergence between the actual output distribution $Distr(a_{t+j})$, and the predicted output distribution $\hat{Distr}(a_{t+j})$. Thus, the problem becomes minimizing: 

\begin{equation}\label{m2_prob}
D_{KL}(Distr^{encoding}(a_{t+j})||\hat{Distr}^{encoding}(a_{t+j}))
\end{equation} where $D_{KL}$ represents the KL divergence. KL-divergence is calculated as per Equation \ref{KL}.

\begin{equation}\label{KL}
KL(p||q) \triangleq \sum_{k=1}^K p_k \log \frac{p_k}{q_k} = \sum_{k=1}^K p_k \log p_k - \sum_{k=1}^K p_k \log q_k
\end{equation} where $q$ is the output probability distribution of our {\Distr2vec} model. $q$ is also an approximation of $p$, the true distribution in the plan trace $Distr(a_{t+j})$. We try to minimize the {\it inclusive} KL divergence \cite{minka-diverg} with the model's target distribution. An advantage of using inclusive KL divergence, is that we avoid computing the derivative of the entropy of $p$ when taking partial derivative of $KL(p||q)$ with respect to model parameters. This is because the values for $p$ (which is $Distr(a_{t+j})$), is a constant with respect to the model parameters. Using this information, we can obtain Equation \ref{KL-m2}.
\begin{eqnarray}\label{KL-m2}
D_{KL}(Distr^{encoding}(a_{t+j})||\hat{Distr}^{encoding}(a_{t+j})) \nonumber\\=Z(Distr(a_{t+j})) -\sum_{k=1}^K c_{t+j}^k \log{p(a_{t+j}^k|h(Distr(a_t)))}
\end{eqnarray} where $Z(Distr(a_{t+j}))=\sum_{k=1}^K c_{t+j}^k \log(c_{t+j}^k)$ is a constant, and $h(Distr(a_t))$ is the embedding computed by multiplying the embedding matrix $W_E$ and the action-distribution input vector $Distr^{encoding}(a_t)=\langle 0...0,c_t^1,0,...,0,c_t^2,0,...0,c_t^K,0... \rangle$ encoded from $Distr(a_t)$, as done in Equation \ref{h-m2}.

\begin{equation}\label{h-m2}
h(Distr(a_t))=W_E \times Distr^{encoding}(a_t)
\end{equation}

\subsection{Combining with Hierarchical Soft-max}
We adopted the hierarchical softmax introduced in \cite{word2vec}, which has been shown to have advantages over the non-hierarchical counterpart in both the accuracy, and computational efficiency\footnote{The first version that tried to improve Word2Vec with hierarchical softmax in \cite{morin2005}, reports that the new model requires less training time, but also results in a degraded accuracy. However, the work \cite{mnih2009} introduces an approach to automatically grow a tree to organize words, for hierarchical softmax, which outperforms non-hierarchical Word2Vec. In the recent work \cite{word2vec}, the advantage is confirmed and it selects the binary Huffman tree as the basic tree data structure.}. For convenience, a brief review of the hierarchical softmax is made in Section \ref{wvhs}.

If we combine Equation \ref{KL-m2} with hierarchical softmax, with distribution input $Distr(a_t)$, we obtain the probability of an action $a_{t+j}^k$ in the target observed action distribution $Distr(a_{t+j})$:

\begin{eqnarray} \label{sm-m2}
p(a_{t+j}^k|h(Distr(a_t))) =\prod_{i=1}^{L(a_{t+j}^k)-1}  \Big\{\sigma(\one(n(a_{t+j}^k,i+1) \\ \nonumber=child(n(a_{t+j}^k,i)))\cdot v_{n(a_{t+j}^k,i)}\cdot h(Distr(a_t)))\Big\}
\end{eqnarray} where $\one(x)$ is an identity function. $L(a_{t+j}^k)$ is the length of path from root to the leaf node, i.e., an action. $v_{n(a_{t+j}^k,i)}$ is the weights vector of $i$th node along the path, and $h$ is the embedding computed using Equation \ref{h-m2}. 

And if we combine Equation \ref{sm-m2} and Equation \ref{KL-m2}, we obtain the error function in Equation \ref{final-m2}. This is the error function as we are trying to minimize the KL divergence of Equation \ref{KL-m2}. 

\begin{eqnarray} \label{final-m2}
E=Z(Distr(a_{t+j})) - \sum_{k=1}^K c_{t+j}^k \sum_{i=1}^{L(a_{t+j}^k)-1} \nonumber \\ \Big\{\log \sigma(\one(.)v_{n(a_{t+j}^k,i)}\cdot h(Distr(a_t)))\Big\}
\end{eqnarray} 

For the detailed derivation of gradient descent with the error computed in Equation \ref{final-m2}, please refer to Section \ref{md-bp}.

\section{Evaluation} \label{evaluation}
	We evaluate the performance of {\Distr2vec} and {\RBM} in {\UDUP} by comparing the plan recognition performance with {\UDUP} and also against a baseline model which we call Naive Model ({\NM}). 

In {\NM}, we feed what the perception module considers the ground truth to {\Word2vec} for training shallow domain models. This perceived ground truth trace is obtained by choosing the most confident (probable) action from each step's distribution. 
We ran our experiments on a machine with a Quad-Core CPU (Intel Xeon 3.4GHz), a 64GB RAM, a GeForce GTX 1080 GPU, and Ubuntu 16.04 OS.
    
\subsection{Dataset for Evaluation}  
We used two datasets: One from real world videos, and the other is a controlled synthetic dataset. In order to sufficiently assess the validity and effectiveness of our  {\Distr2vec} approach, we need to test input data with different types of distributions. Thus we generated a synthetic dataset, as it would allow us to vary the parameters of the distribution for maximally evaluating the approach. We also collected a dataset based on real world videos, to test how effective our model is with real world data. 

We created a synthetic dataset of action distribution sequences, from the 50 Salads Dataset \cite{Stein_2013}. The 50 Salads dataset contains segmented videos, and each segment has a low-level activity label, for example ``add\_oil\_prep". In this work we use the 54 ground-truth, low-level activity sequences, $a^*_{1:T}$ (T is sequence length).

The traces from the 50 Salads Dataset, have ground truth actions (no distribution over the actions). We synthesize a distribution per action step, by adding actions to each step and assigning a probability distribution over the actions. We add $K-1$ additional actions to each observation and assign a probability distribution such that the ground truth has the highest confidence (probability). 
    
In order to generate the $K-1$ additional actions, we search for the $K-1$ most similar actions by using a {\Word2vec} model that is pretrained on Google News corpus \footnote{\url{https://drive.google.com/file/d/0B7XkCwpI5KDYNlNUTTlSS21pQmM/edit}}. We use the semantic correlation of actions in this model, to simulate their visual correlation. As for assigning confidence values for each of the $K-1$ additional actions, we followed the Equation \ref{fakeConf}.

\begin{equation} \label{fakeConf}
c(a^k_t) = \frac{s_k}{1+w_{entropy}+\sum_{i=1}^{K-1} s_i}
\end{equation} where $a^k_t$ is the $k^{th}$ additional action in a distribution at step $t$. We search for $a^k_t$ in the {\Word2vec} model based on the ground truth action $a^*_t$. $s_k$ is the similarity between $a^*_t$, and $a^k_t$, which can be computed by using the pretrained {\Word2vec} of gensim library \cite{rehurek_lrec}. These similarity values are used as in the Equation \ref{fakeConf}.The $+1$ in the denominator denotes the similarity between the ground-truth action and itself. We also use a parameter $w_{entropy}$  which we can use to increase or decrease probability differences between actions. Thus it could be used to adjust the entropy of each action distribution. We set up $w_{entropy}$ for the computation of each $c(a^k_t)$ to either zero or one, to slightly increase the variance of confidences at each step. 
The confidence for ground-truth action $a^*_t$ is initially the highest and is equal to Equation \ref{gnd_a}.
\begin{equation} \label{gnd_a}
c(a^*_t) = 1 - \sum_{k=1}^{K-1} c(a^k_t)
\end{equation} 
where $c(a^k_t)$ is the confidence of one of the $K-1$ additional actions in the distribution at a step. 

Thus far, the synthetic data generated will have the ground truth as the action with the highest confidence. We would like to vary this, and simulate perception errors. We define a perception error (PE) as when the action $a_t$ that has the highest confidence in $Distr(a_t)$ does not match the ground-truth action $a^*_t$. The perception error {\it rate} (PER) is the percentage of action distributions that have perception errors. We simulate the PE in a particular action distribution by exchanging the highest confident action with another action in that distribution. A specified PER is achieved by simulating PE in a proportion of the action distributions in each plan trace. The action distributions in which we simulate PE are randomly chosen. In this way, data of different distribution types is generated for testing the effectiveness of our model.

\begin{figure*}[t] 
  \centering
  \includegraphics[width=14cm]{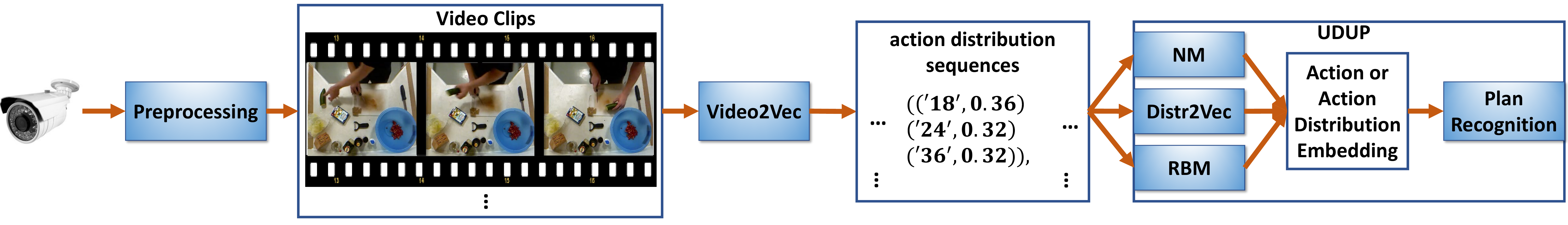}
  \caption{Flowchart illustrating how we evaluate three action affinity models for plan recognition, on real world videos.}
  \label{fig:flowchart}
\end{figure*}

As for using real world videos to generate action distribution sequences, we also used the aforementioned 50 Salads Dataset \cite{Stein_2013}. The procedure is illustrated in Figure \ref{fig:flowchart}. First, we converted all videos into video clips, and each clip matches a ground-truth action. Secondly, we applied a video processing model, Video2Vec \cite{hu2016video2vec}, on those video clips, to output action distribution sequences. For each clip it outputs a distribution over all actions. We split the video clips of each action using 9:1 ratio, for training and testing respectively. Then we train the Video2Vec\cite{hu2016video2vec} for 450 epochs. The visual classification accuracy on these video clips is about 95\% on training dataset, and about 65\% on testing dataset. By using the trained Video2Vec, we obtain a low PER (8\%) plan corpora. We then re-split the video clips using 7:3 ratio, for training and testing respectively, and train the second Video2Vec \cite{hu2016video2vec} with fewer epochs (i.e. 100 epochs). The classification accuracy of visual model is about 42\% on training dataset, and about 17\%  on testing dataset. We obtain a higher PER (79\%) plan corpora by using the second Video2Vec. Finally, we concatenate distribution sequences together for each video to make plan corpora that correspond to real-world videos. 

\begin{figure*}[t] 
  \centering
  \includegraphics[width=15cm]{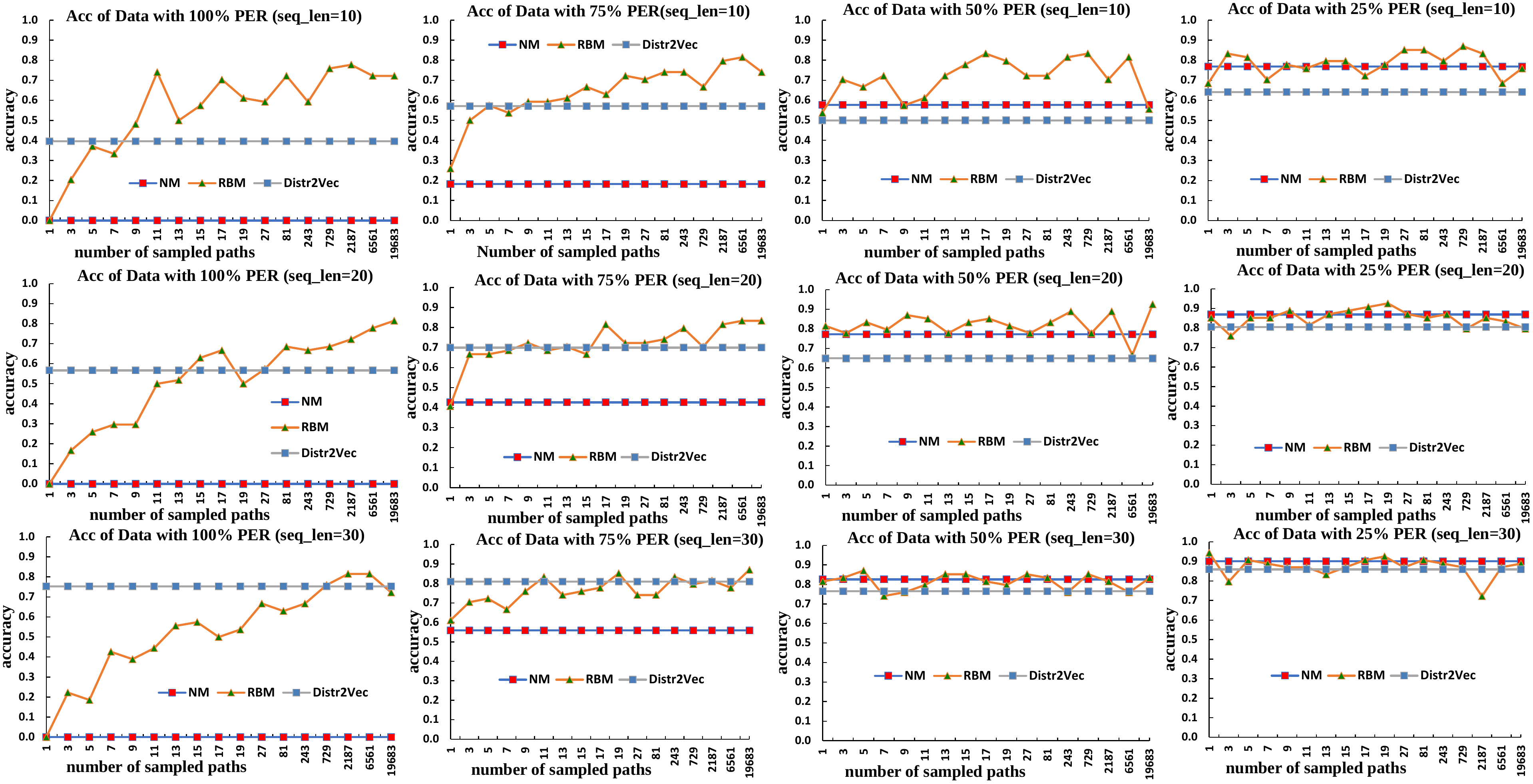}
  \caption{This figure demonstrates the accuracy of three models on synthetic data of different lengths and \textbf{PERs}, with respect to varying number of samples.}
  \label{fig:eval1}
\end{figure*}
\begin{figure*}[t] 
  \centering
  \includegraphics[width=12cm]{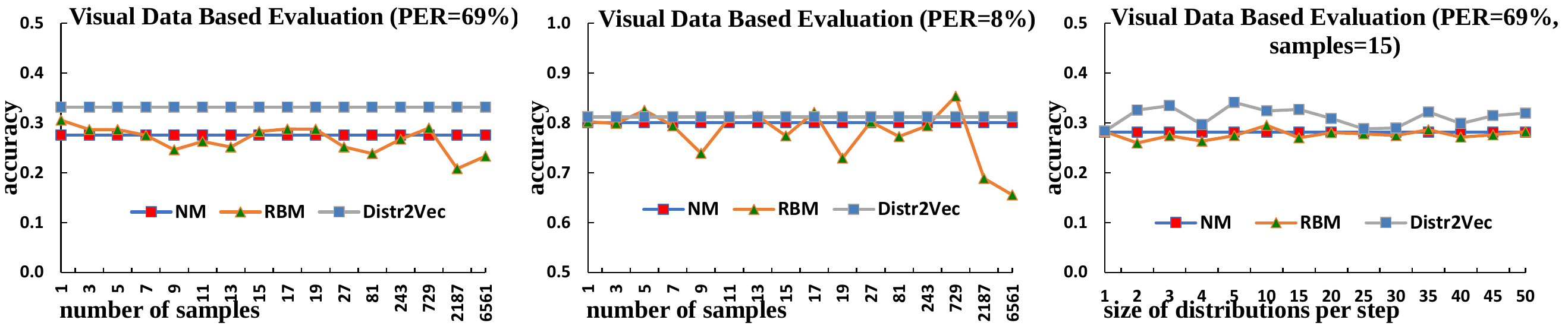}
  \caption{Results of evaluating the {\NM}, {\RBM}, and {\Distr2vec} on videos, which demonstrate the effectiveness of {\Distr2vec}.}
  \label{fig:real-world}
\end{figure*}

\subsection{Testing Methodology}  
	We evaluate different shallow domain models as used in {\UDUP} using 6-fold cross validation. We train the shallow domain models (either in the form of {\Distr2vec} or {\Word2vec}), and then measure the {\UDUP} model's performance when using each of the trained planning domain models. For each test, we randomly remove some distributions at some positions. Then we compare the performance by the average accuracy. We define the accuracy in the same way as in the {\DUP} model work \cite{dup}, as follows:

\begin{equation} \label{acc}
acc=\frac{1}{Z} \sum_{i=1}^Z \frac{\#\langle CorrectSuggestions\rangle_i}{K_i}
\end{equation} 
Where $Z$ is the number of traces in testing set, and $K_i$ is the number of missing positions for the trace $i$. From each of the traces in $Z$, we add up the ratio of the number of correct suggestions to the number of missing actions.

In our experiments on the synthetic dataset, we analyze the effect of the parameters of $w_{entropy}$, PER, and the length of observation sequences. We analyze the effect of sampled path numbers, and distribution sizes, in experiments on distribution sequence dataset extracted from real world videos. 

More specifically, in accuracy evaluation on synthetic dataset, we evaluate the data under two categories: 1) Fixed entropy but varying PERs between (25\%, 50\%, 75\%, and 100\%); 2) Zero PER but test with both high entropy (all confidence values each time step are uniformly distributed) and low entropy. We set PER to zero in order to  evaluate with only the influence of entropy. For low entropy, we set the confidence for $a^*_t$ to 0.9, and the two simulated actions to 0.05. And on real world dataset (consists of whole distribution sequences with varying lengths), we also evaluate the data under two categories: 1) Fixed sequence length (30) but varying number of samples; 2) Fixed number of samples (15) but varying sizes of distributions.

We also look at the effect of the number of sampled paths on the training time, on the synthetic dataset. Please note that in results where the number of samples are varied, it only affects the {\RBM} model. The other models are unaffected by resampling and thus only have their average value plotted as a line. 

All of our experiments have the following hyper\-parameters (thus fair to all models): the number of missing actions is one in synthetic evaluation and 10\% in video based evaluation, the context window  size is one, the number of recommendations for each missing action is three, and the number of threads when running the experiment is eight. For all testing scenarios, the accuracy is evaluated by having the model predict the action at a randomly picked position. The accuracy is calculated using Equation \ref{acc}.

\subsection{Analysis with regard to Accuracy (Synthetic Dataset)}

We show our results in the Figure \ref{fig:eval1} for traces of length 10,20 and 30, with varying PER and fixed entropy. We show the results of varying entropy and fixed PER in the supplemental material, Figure 1. For 100\% PER, the prediction of {\Distr2vec} gets better for longer training sequences. This matches our expectation because more training data is available for learning better embeddings. As the trace length increases, the {\RBM} model requires more samples from the distribution sequence to match or outperforms the {\Distr2vec} model. For traces of length 30, it takes significantly more sampling steps to match the performance of {\Distr2vec} model. We interpret more sampling to mean more of the distributional data is captured in the sampled traces. This also means that {\Distr2vec} model better captures the information in the distributional data with longer sequences (and thus more training data) since it takes more samples for the {\RBM} approach to match the performance. For other lower PER cases as well, a similar pattern is seen as in the 100\% PER case. With higher trace lengths, the {\Distr2vec} model performs better and the {\RBM} model needs more samples to match performance for larger trace length data.

As for the {\NM}, accuracy for 100\% PER across all lengths is 0\%. This matches our expectations because the training data for the {\NM} model does not have the ground-truth action in any step for 100\% PER. Therefore, the {\NM} would not capture any relevant information. For lower PER cases the {\NM} performance is higher with lower perception error. This matches our expectations since the {\NM} gets more of the ground truth information from highest-probability sampling of low PER traces. 

\begin{figure}[thpb] 
  \centering
  \includegraphics[width=8cm]{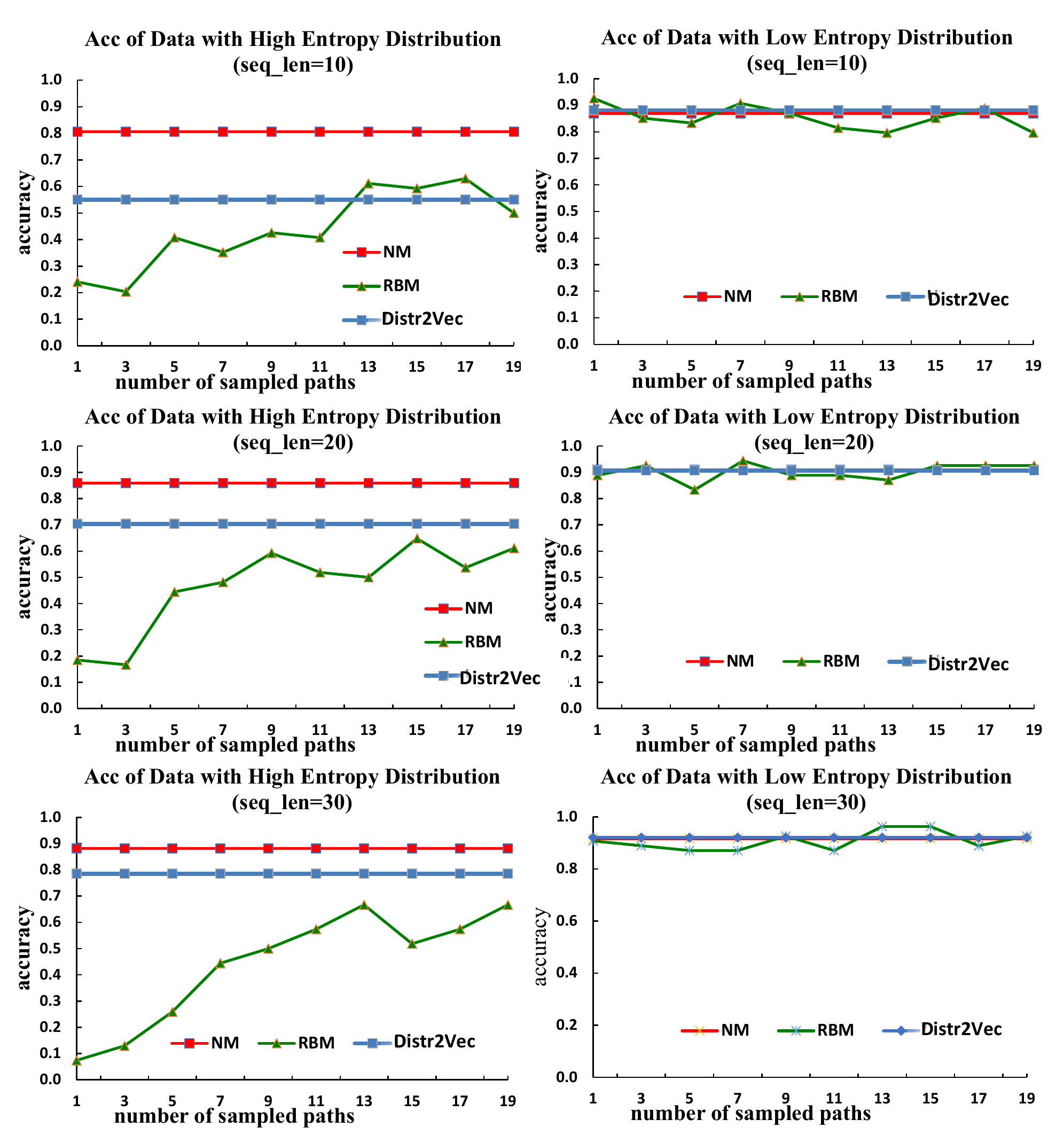}
  \caption{This figure demonstrates the accuracy of three models on synthetic data of different lengths and \textbf{entropies}, with respect to varying number of samples.}
  \label{fig:eval2}
\end{figure}

Lastly, for data with high and low entropies, we set PER to zero. The experiments results is shown in Figure. \ref{fig:eval2}. As expected, {\NM} has the best accuracy since it get the ground-truth plan trace when PER is zero. We also observe that, the {\Distr2vec} predicts missing actions with comparable accuracy for low entropy. At high entropy the performance of {\Distr2vec} is expectedly lower than the performance at low entropy, and it gets appreciably better with longer training data. We also note that {\Distr2vec} is clearly better than the {\RBM} for high entropy. This makes sense because resampling for the {\RBM} can result in more incorrect plan traces for training, when entropy is high. It is worth noting that the naive {\NM} outperforms {\Distr2vec} for high entropy due to an artifact of the way we simulated high entropy. The ground truth was still the most probable action, and so the {\NM} learned its shallow domain model from the ground truth. That is why {\NM} has such a high accuracy for high entropy cases as well.

\subsection{Analysis with regard to Accuracy (Video Based Dataset)}
The results are shown in Figure \ref{fig:real-world}: the left two plots show how number of samples influence the accuracy, when PER is high (69\%) and low (8\%) (setting distribution size to 5). The rightmost plot shows how sizes of distributions per step influence the accuracy (setting PER to 69\%). We can observe that, when PER is high, the plan recognizer with {\Distr2vec} performs the best, whereas when PER is low, all models perform comparably well. This is consistent to results of synthetic data evaluation. We can also observe that, overall the distribution size is not a key factor for deciding model performance.

\subsection{Analysis with regard to Training Time}
\begin{figure*}[t] 
  \centering
  \includegraphics[width=12cm]{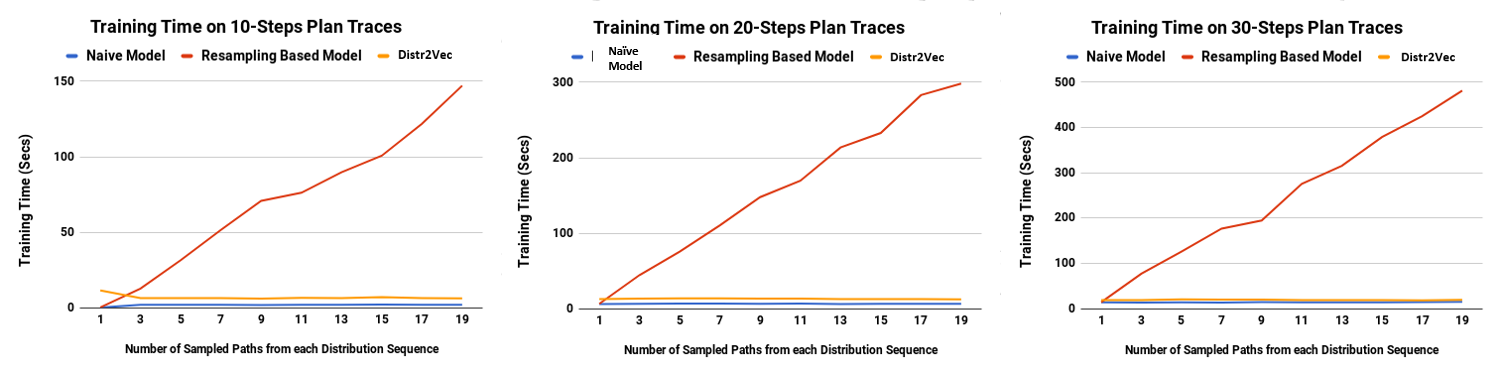}
  \caption{This figure shows the comparison of training time.}
  \label{fig:training_time}
\end{figure*}

For training time analysis, it is pertinent to note that all models use the Python {\Word2vec} functions from the {\it gensim} library \cite{rehurek_lrec}. For training the {\RBM}, we varied the number of sampled paths per plan trace. We compared the training time across all models and the data is presented in Figure \ref{fig:training_time}. We observe that the training time of {\RBM} increases linearly with the number of samples per trace. This is easily explained by the fact that increasing the number of samples, is increasing the training data, and thus takes more time to train. In contrast, the training time for {\Distr2vec} and {\NM} stay constant as there is no sampling step.

\section{Related work} \label{related-work-background}
To the best of our knowledge, we are the first to learn embeddings and shallow planning domain models from a sequence of distributions. In a more general sense, there are some works that address the problem handling uncertain data in classification tasks \cite{qin2009rule,ren2009naive,ge2010unn}. 
The closest work to ours is \cite{ge2010unn} which did take in uncertain training data and modeled it as a Gaussian distribution for processing. However, they used a linear perceptron for classification. They do not learn embeddings for the input distribution to measure affinities as our model does, rather they only classify the uncertain data into categories. In comparison, we extend the concept of word embeddings to learn embeddings for uncertain data distributions. These embeddings allow us to quantify affinities and relationships between the input distributions in a sequence learning problem, which is very different to classification.

From the plan recognition point of view, our work addresses the problem of learning a shallow planning domain model for plan recognition. In the planning literature, there has been a long history of leveraging models to recognize plans. In \cite{sohrabi2016plan} and  \cite{geffner-ramirez}, solving a plan recognition problem is transformed to solving a planning problem. These two plan recognition works assume that a planning domain model is given, whose construction strictly follows certain rules or syntaxes, like Planning Domain Definition Language \cite{mcdermott1998pddl}. As discussed in \cite{dup} and \cite{rao-model-lite}, such models are termed as full models. \cite{rao-model-lite} and \cite{dup} also provide a detailed discussion with respect to differences between full model, approximated model, and shallow model, that we could use to provide planning support. Our UDUP framework belongs to shallow model based plan recognition. A shallow planning model is learned and could be flexibly represented in various formats. 
Both \cite{dup} and \cite{zhuo2017human} apply {\Word2vec} to learn shallow planning models from plan corpora. In contrast to these works, our UDUP with {\RBM} uses {\Word2vec} in a specific way (importance resampling) to handle uncertain action observations, and our UDUP with {\Distr2vec} directly handles uncertain observations by using our proposed KL-divergence based loss function.

\section{Conclusion} \label{conclusion}
We introduced our {\Distr2vec} and {\RBM} model, that learns embeddings for distributions. We then applied them to do plan recognition with our {\UDUP}. {\UDUP} can learn an shallow planning domain model (based on the learned embeddings), from plan traces of distributions over actions. The learned shallow (or shallow) domain model would then be used by {\UDUP} to search for the most probable actions for missing positions in an incomplete plan. Unlike {\DUP}, {\UDUP} can be trained on traces of observed action distributions, and thus can handle uncertainty in the input.

We evaluate our models ({\RBM} and {\Distr2vec}) in the context of plan recognition in {\UDUP} by evaluating it against {\UDUP} that uses a baseline  domain model trained with normal {\Word2vec}. We evaluated all models on two datasets. One is a synthetic dataset that we could use to evaluate with different trace lengths, PERs, and entropies. The other is a video based dataset, which allows us to assess the real-world value of our models. From the experimental results on both datasets, we can conclude that when there is a higher PER, {\UDUP} with {\Distr2vec} outperforms other models. When the PER is lower, and sequence length is longer, all models perform comparably well. {\Distr2vec} model's performance improves markedly with more and longer training sequences. Additionally, for synthetic dataset experiments, when the entropy of the data is higher, the {\Distr2vec} still produces appreciable-quality domain models which keeps up the accuracy of the {\UDUP}. Another benefit of using {\Distr2vec} with {\UDUP} is that the training time is shorter as compared to the {\RBM} model when we sample more traces, and almost equal to the {\NM} baseline model. The training time of {\RBM} increases linearly with the number of samples taken per plan trace. Comparing the training time of {\Distr2vec} model with {\RBM} is more appropriate because both models factor in the uncertainty in the training data. However, {\NM} discards all information about the uncertainty in the data, and thus losses information.

\section{Appendix} \label{Appendix}
\subsection{Word2Vec and Hierarchical Softmax} \label{wvhs}
In this section we make a brief review about how the Skip-gram {\Word2vec} works, based on \cite{word2vec}, and how it could be used for plan recognition as introduced in \cite{dup}. To be consistent with the whole paper, we use the term ``action'' and treat it an equivalence to the term ``word'' in other papers which introduces {\Word2vec}.

Given a corpora which contains action (word) sequences, a Skip-gram model is trained by maximizing the average log probability:

\begin{eqnarray} \label{wv1}
\frac{1}{T} \sum_{t=1}^T \sum_{-\mathcal{W}\leq j \leq \mathcal{W}, j \neq 0} \log p(a_{t+j}|a_t)
\end{eqnarray} where $T$ is the length of a sequence, $\mathcal{W}$ is the context window size, $a_t$ is fed as the model input, $a_I$, and $a_{t+j}$ is used as the target action (word), $a_O$. The probability $p(a_O|a_I)$ is computed as:

\begin{eqnarray} \label{wv2}
p(a_O|a_I) =\frac{exp({v'}_{a_O}^T v_{a_I})}{\sum^A_{a=1} exp({v'}_a^T v_{a_I})}
\end{eqnarray} which is essentially a softmax function. $A$ denotes a vocabulary of all possible actions. Other symbols follows the definition in Equation \ref{wv1}. We can use this equation that computes $p(a_O|a_I)$ to compute $p(a_{t+j}|a_t)$ in Equation \ref{wv1}.

The $p(a_{t+j}|a_t)$ in the {\Word2vec} with hierarchical softmax is calculated in the following manner. The output layer's weight matrix of the regular {\Word2vec} is replaced by a binary tree whose leaf nodes are words in the trained vocabulary. Every node on the path from the root node until the leaf node has a vector, excluding the leaf node. The input action $a_i$ is converted into an embedding $h$ which is the input into the binary tree component. The probability of this input vector $h$ that could go to a particular leaf node is calculated by following the path from the root node to the target leaf node, using the following formula:

\begin{eqnarray}\label{hs-prediction}
p(a_{t+j}|a_t)=\prod_{i=1}^{L(a_{t+j})-1}\Big\{\sigma(\one(n(a_{t+j},i+1)=\nonumber \\child(n(a_{t+j},i)))\cdot v_{n(a_{t+j},i)}\cdot h)\Big\}, 
\end{eqnarray} 
	where $\one(x)$ is a function that returns 1 if the next node on the path to the target leaf node is on the left of the current node, and -1 if the next node is to the right. $L(a_{t+j})$ is the length of path from root to the leaf node $a_{t+j}$, $v_{n(a_{t+j},i)}$ is the vector of the $i$th node along the path. $h$ is the embedding obtained by multiplying the embedding matrix and vector of the input action $a_t$. $h$ is the vector that represents the input action in the embedding space.
	In order to maximize the probability, the vectors of the intermediary nodes are updated with each training sample which has the target action $a_t$ and an action in its context $a_{t+j}$.

\subsection{Derivation of Gradient Update Equations} \label{md-bp}
\begin{figure}[!ht]
\begin{center}
	\centerline{\includegraphics[width=0.4\textwidth]{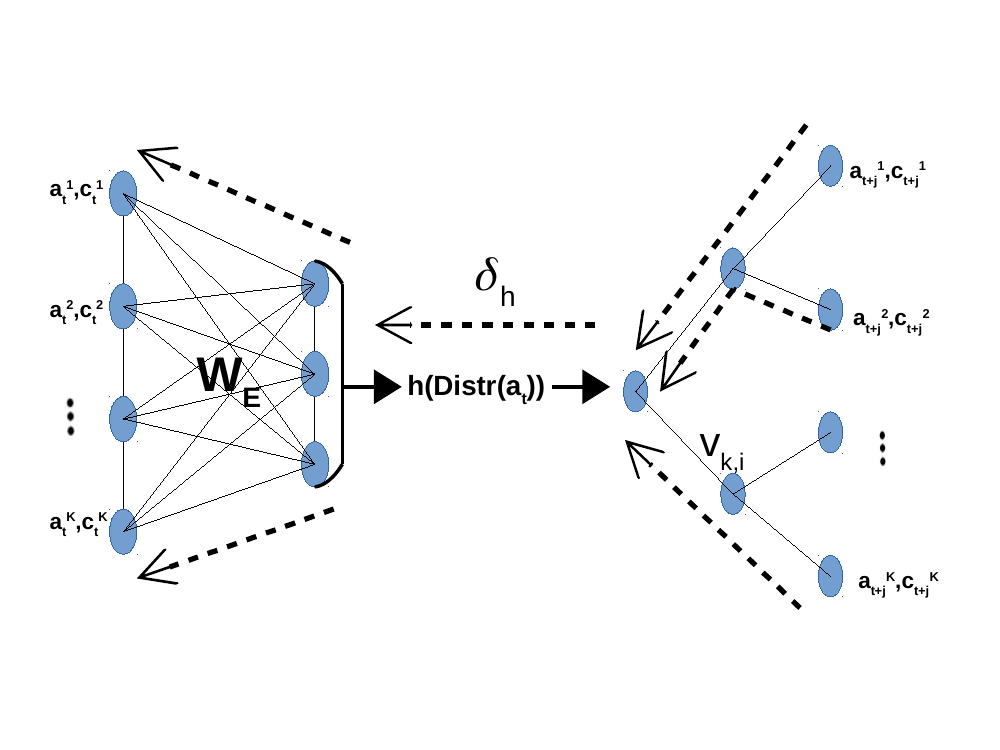}}
	\caption{Details of {\Distr2vec} model for Error Propagation.}
	\label{backprop_error_img}
\end{center}
\end{figure}


We back-propagate the error $E$ as shown in Figure \ref{backprop_error_img}, where the dashed lines show the path of back-propagation. For this derivation, we shorten $h(Distr(a_t))$ as $h$. We start by computing the derivative of $E$ with respect to $(v_{n(a_{t+j}^k,i)}*h)$ as in Equation \ref{pepvh}. We also shorten our notation of $v_{n(a_{t+j}^k,i)}$ as $v_{k,i}$, which is the vector of $i$-th node in the tree path to the leaf node of the $k$-th action in the vocabulary as illustrated in Figure \ref{backprop_error_img}.

{\scriptsize
\begin{eqnarray} \label{pepvh}
\frac{\partial E}{\partial v_{k,i} h} \nonumber = -c^k_{t+j}\frac{\frac{\partial \sigma(\one(.)v_{k,i}h)}{\partial v_{k,i}h}}{\sigma(\one(.)v_{k,i}h)}
\end{eqnarray}
\begin{eqnarray}
=-c^k_{t+j}\frac{\sigma(\one(.)v_{k,i}h)(1-\sigma(\one(.)v_{k,i}h) \one(.)}{\sigma(\one(.)v_{k,i}h)} \nonumber \\ =c^k_{t+j} (\sigma(\one(.)v_{k,i}h)-1) \one(.) = \begin{cases} c^k_{t+j} (\sigma(\one(.)v_{k,i}h)-1) ,(\one(.)=1)\\ c^k_{t+j} (\sigma(\one(.)v_{k,i}h) ,(\one(.)=-1) \end{cases} \nonumber \\ = c^k_{t+j} (\sigma(v_{k,i}h) - t_i)
\end{eqnarray}} where $t_i=1$ if $\one(.)=1$ and $t_i=0$ if $\one(.)=-1$. Recall that $\one(.)$ is the identity function defined in Equation \ref{sm-m2}. 

Then we calculate the derivative with respect to each $v_{k,i}$ along the path to a specific action, at the time step $t+j$:
\begin{eqnarray}
\frac{\partial E}{\partial v_{k,i}}=\frac{\partial E}{\partial v_{k,i}h} \frac{\partial v_{k,i}h}{\partial v_{k,i}}=c^k_{t+j} (\sigma(v_{k,i}h) - t_i)h
\end{eqnarray}

With this, we update each $v_{k,i}$ as follows:
\begin{eqnarray}
v_{k,i} = v_{k,i} - \alpha \frac{\partial E}{\partial v_{k,i}}
\end{eqnarray} Note that each node's vector $v_{k,i}$ could get updated more than once, as each node could be on the path to more than one action as show in the right side of Figure \ref{backprop_error_img}. 

Then we compute the back-propagated error $\delta_h$ by substituting Equation \ref{pepvh} as follows:
\begin{eqnarray} \label{delta_h1}
\delta_h = \frac{\partial E}{\partial h} = \sum_{k=1}^K \sum_{i=1}^{L(a_{t+j}^k)-1} \frac{\partial E}{\partial v_{k,i}h} \frac{\partial v_{k,i}h}{\partial h} \nonumber \\= \sum_{k=1}^K c_{t+j}^k \sum_{i=1}^{L(a_{t+j}^k)-1} \sigma(v_{k,i}h) - t_i)v_{k,i} 
\end{eqnarray} We can understand this equation by imagining that there are multiple channels coming back from each leaf node, passing thorough a sequence of child nodes in the hierarchical softmax tree, towards the output of the embedding matrix $W_E$. So doing the back-propagation means summing errors of these channels together, and that is why $v_{k,i}$ could get updated more than once.

We derive the Equation \ref{delta_h1} leveraging Equation \ref{pepvh}. However, we could also go directly from the original error function (Equation \ref{final-m2}). Thus here we provide another derivation of $\delta_h$ which is equally valid:

\begin{gather}
\delta_h = \frac{\partial E}{\partial h} = \frac{\partial [-\sum_{k=1}^K c_{t+j}^k \sum_{i=1}^{L(a_{t+j}^k)-1} \log \sigma(\one(.)v_{n(a_{t+j}^k,i)}\cdot h)]}{\partial h} \nonumber \\ = -\sum_{k=1}^K c_{t+j}^k [\sum_{i=1}^{L(a_{t+j}^k)-1} \frac{\partial \log \sigma(\one(.)v_{n(a_{t+j}^k,i)}\cdot h)}{\partial h}] \nonumber \\ = -\sum_{k=1}^K c_{t+j}^k [\sum_{i=1}^{L(a_{t+j}^k)-1} (\sigma(v_{n(a_{t+j}^k,i)}h) - t_i)v_{n(a_{t+j}^k,i)}]
\end{gather} where $\frac{\partial \log \sigma(\one(.)v_{n(a_{t+j}^k,i)}\cdot h)}{\partial h}$ has already been derived as a part of Equation \ref{pepvh}.

Finally, we update the weights in the embedding matrix $W_E$:
\begin{eqnarray}
\frac{\partial E}{\partial W_E} = \frac{\partial E}{\partial h} \frac{\partial h}{\partial W_E} = \delta_h v_a^t
\end{eqnarray} where $v_a^t = \langle c_t^1,c_t^2,...,c_t^K \rangle$ is the confidence values from $Distr(a_t)$, and $\frac{\partial E}{\partial W_E}$ can be used to update the values in $W_E$ as follows:

\begin{equation}
W_E = W_E - \alpha*\frac{\partial E}{\partial W_E}
\end{equation}

\bibliographystyle{aaai.bst}  
\bibliography{main.bbl}  


\bibliographystyle{ACM-Reference-Format}  
\bibliography{aaai17.bib}  

\begin{thebibliography}{}

\bibitem[\protect\citeauthoryear{{ R}eh{ u}{ r}ek and
  Sojka}{2010}]{rehurek_lrec}
{ R}eh{ u}{ r}ek, R., and Sojka, P.
\newblock 2010.
\newblock {Software Framework for Topic Modelling with Large Corpora}.
\newblock In {\em {Proceedings of the LREC 2010 Workshop on New Challenges for
  NLP Frameworks}},  45--50.
\newblock Valletta, Malta: ELRA.

\bibitem[\protect\citeauthoryear{Ge, Xia, and Nadungodage}{2010}]{ge2010unn}
Ge, J.; Xia, Y.; and Nadungodage, C.
\newblock 2010.
\newblock Unn: a neural network for uncertain data classification.
\newblock {\em Advances in Knowledge Discovery and Data Mining}  449--460.

\bibitem[\protect\citeauthoryear{Hu, Li, and Li}{2016}]{hu2016video2vec}
Hu, S.-H.; Li, Y.; and Li, B.
\newblock 2016.
\newblock Video2vec: Learning semantic spatial-temporal embeddings for video
  representation.

\bibitem[\protect\citeauthoryear{Kambhampati}{2007}]{rao-model-lite}
Kambhampati, S.
\newblock 2007.
\newblock Model-lite planning for the web age masses: The challenges of
  planning with incomplete and evolving domain models.
\newblock In {\em Proceedings of the Twenty-Second {AAAI} Conference on
  Artificial Intelligence},  1601--1605.

\bibitem[\protect\citeauthoryear{Lipowski and
  Lipowska}{2012}]{lipowski2012roulette}
Lipowski, A., and Lipowska, D.
\newblock 2012.
\newblock Roulette-wheel selection via stochastic acceptance.
\newblock {\em Physica A: Statistical Mechanics and its Applications}
  391(6):2193--2196.

\bibitem[\protect\citeauthoryear{McDermott \bgroup et al\mbox.\egroup
  }{1998}]{mcdermott1998pddl}
McDermott, D.; Ghallab, M.; Howe, A.; Knoblock, C.; Ram, A.; Veloso, M.; Weld,
  D.; and Wilkins, D.
\newblock 1998.
\newblock Pddl-the planning domain definition language.

\bibitem[\protect\citeauthoryear{Mikolov \bgroup et al\mbox.\egroup
  }{2013}]{word2vec}
Mikolov, T.; Sutskever, I.; Chen, K.; Corrado, G.~S.; and Dean, J.
\newblock 2013.
\newblock Distributed representations of words and phrases and their
  compositionality.
\newblock In {\em NIPS},  3111--3119.

\bibitem[\protect\citeauthoryear{Minka and others}{2005}]{minka-diverg}
Minka, T., et~al.
\newblock 2005.
\newblock Divergence measures and message passing.
\newblock Technical report, Technical report, Microsoft Research.

\bibitem[\protect\citeauthoryear{Mnih and Hinton}{2009}]{mnih2009}
Mnih, A., and Hinton, G.~E.
\newblock 2009.
\newblock A scalable hierarchical distributed language model.
\newblock In {\em Advances in neural information processing systems},
  1081--1088.

\bibitem[\protect\citeauthoryear{Morin and Bengio}{2005}]{morin2005}
Morin, F., and Bengio, Y.
\newblock 2005.
\newblock Hierarchical probabilistic neural network language model.
\newblock In {\em Aistats}, volume~5,  246--252.
\newblock Citeseer.

\bibitem[\protect\citeauthoryear{Qin \bgroup et al\mbox.\egroup
  }{2009}]{qin2009rule}
Qin, B.; Xia, Y.; Prabhakar, S.; and Tu, Y.
\newblock 2009.
\newblock A rule-based classification algorithm for uncertain data.
\newblock In {\em Data Engineering, 2009. ICDE'09. IEEE 25th International
  Conference on},  1633--1640.
\newblock IEEE.

\bibitem[\protect\citeauthoryear{Ram{\'{\i}}rez and
  Geffner}{2009}]{geffner-ramirez}
Ram{\'{\i}}rez, M., and Geffner, H.
\newblock 2009.
\newblock Plan recognition as planning.
\newblock In {\em {IJCAI} 2009, Proceedings of the 21st International Joint
  Conference on Artificial Intelligence, Pasadena, California, USA, July 11-17,
  2009},  1778--1783.

\bibitem[\protect\citeauthoryear{Ren \bgroup et al\mbox.\egroup
  }{2009}]{ren2009naive}
Ren, J.; Lee, S.~D.; Chen, X.; Kao, B.; Cheng, R.; and Cheung, D.
\newblock 2009.
\newblock Naive bayes classification of uncertain data.
\newblock In {\em Data Mining, 2009. ICDM'09. Ninth IEEE International
  Conference on},  944--949.
\newblock IEEE.

\bibitem[\protect\citeauthoryear{Sohrabi, Riabov, and
  Udrea}{2016}]{sohrabi2016plan}
Sohrabi, S.; Riabov, A.~V.; and Udrea, O.
\newblock 2016.
\newblock Plan recognition as planning revisited.
\newblock In {\em IJCAI},  3258--3264.

\bibitem[\protect\citeauthoryear{Stein and McKenna}{2013}]{Stein_2013}
Stein, S., and McKenna, S.~J.
\newblock 2013.
\newblock Combining embedded accelerometers with computer vision for
  recognizing food preparation activities.
\newblock In {\em Proceedings of the 2013 ACM International Joint Conference on
  Pervasive and Ubiquitous Computing (UbiComp 2013), Zurich, Switzerland}.
\newblock ACM.

\bibitem[\protect\citeauthoryear{Tian, Zhuo, and Kambhampati}{2016}]{dup}
Tian, X.; Zhuo, H.~H.; and Kambhampati, S.
\newblock 2016.
\newblock Discovering underlying plans based on distributed representations of
  actions.
\newblock In {\em Proceedings of the 2016 International Conference on
  Autonomous Agents \& Multiagent Systems},  1135--1143.
\newblock International Foundation for Autonomous Agents and Multiagent
  Systems.

\bibitem[\protect\citeauthoryear{Zhuo}{2017}]{zhuo2017human}
Zhuo, H.~H.
\newblock 2017.
\newblock Human-aware plan recognition.
\newblock In {\em AAAI},  3686--3693.

\end{thebibliography}

\end{document}